\documentclass{article}
\usepackage{fancyvrb}
\usepackage{ijcai25}
\usepackage{times}
\usepackage{soul}
\usepackage{url}
\usepackage[hidelinks]{hyperref}
\usepackage[utf8]{inputenc}
\usepackage[small]{caption}
\usepackage{graphicx}
\usepackage{amsmath}
\usepackage{amsthm}
\usepackage[numbers]{natbib}
\usepackage{booktabs}
\usepackage{algorithm}
\usepackage{algorithmic}
\usepackage[switch]{lineno}
\usepackage{longtable}
\usepackage{subcaption}
\usepackage{multirow}
\usepackage{colortbl}
\usepackage[table]{xcolor}
\usepackage{float}
\usepackage{listings}
\usepackage{xcolor}

\title{Advanced Financial Reasoning at Scale: A Comprehensive Evaluation of Large Language Models on CFA Level III}

\author{
Pranam Shetty$^1$
\and
Abhisek Upadhayaya$^3$
\and
Parth Mitesh Shah$^2$
\and
Shilpi Nayak$^2$
\and
Anna Joo Fee$^2$
\and
Srikanth Jagabathula$^3$
\and
\affiliations
$^1$Rochester Institute of Technology,
$^2$GoodFin, Inc,
$^3$New York University\\
\emails
ps9960@rit.edu,
\{parth, shilpi, anna\}@goodfin.com,
au2216@nyu.edu, sjagabat@stern.nyu.edu
}

\begin{document}

\maketitle

\begin{abstract}
As financial institutions increasingly adopt Large Language Models (LLMs), rigorous domain-specific evaluation becomes critical for responsible deployment. For advanced financial reasoning, the Chartered Financial Analyst (CFA) Level III exam is widely considered the gold standard. In this paper, we present a comprehensive benchmark evaluating 23 state-of-the-art LLMs on mock CFA Level III exams, which require answering challenging multiple choice and essay questions. We evaluate reasoning and non-reasoning models, both proprietary and open source, using three prompting strategies: zero-shot, chain-of-thought, and self-discover. 
We find that frontier reasoning models, such as o4-mini, Gemini 2.5 Pro, and Claude Opus 4, using chain-of-thought prompting demonstrate strong capabilities, successfully passing the mock Level III exams. While there is little to separate the frontier models on multiple choice questions, only a few models excel at the complex essay questions, which require analysis, synthesis, and strategic thinking. These results demonstrate significant progress in the financial reasoning capabilities of LLMs, which previously~\cite{mahfouz2024state} could clear Level I and Level II exams but struggled with the Level III exam, particularly the essay questions. We also benchmark the models on cost and latency, providing crucial guidance for practitioners on model selection. Our analysis highlights ongoing challenges in cost-effective deployment and underscores the need for nuanced interpretation of performance against professional benchmarks.
\footnote{Accepted at FinLLM@IJCAI 2025}
\end{abstract}

\section{Introduction}

Large Language Models (LLMs) are rapidly transforming the financial services industry, with growing adoption for critical applications including research, education, trading, and client advisory services~\cite{chen2024survey, de2025multi}. As the use of such models proliferates, it is important to ensure that they meet the strict standards necessary for application in high-stakes settings. While the LLMs are demonstrating improving proficiency on general reasoning tasks, a critical question arises: can they also handle specialized, high-stakes analytical reasoning required for professional financial decision-making?~\cite{benton2024sabotage, callanan2023can}

To asses the capabilities of LLMs in highly specialized domains such as finance, we need professionally relevant benchmarks that go beyond simple recall and assess complex cognitive skills. Popular benchmarks \cite{islam2023financebench, wu2023bloomberggpt, maia201818}
tend to test general-purpose quantitative or reasoning skills but are not designed to capture nuanced reasoning required for financial analysis, involving synthesis of quantitative data, regulatory knowledge, and market dynamics to make investment decisions~\cite{mahfouz2024state}.


The Chartered Financial Analyst (CFA) Level III exam represents an ideal evaluation framework, being the culminating assessment for investment management professionals. Level III is the final CFA exam, taken after the candidates have passed Levels I and II, and is centered portfolio management and wealth planning, assessing candidates' ability to implement financial knowledge in real-world scenarios~\cite{cfa_level3_exam, charlton1999cfa}. It uses a mixed format, comprising 11 item set or multiple choice questions (MCQs) and 11 essay or constructed response questions. Each question set begins with a narrative scenario and data (the vignette), which must be carefully analyzed to answer the questions. This dual format comprehensively tests higher-order cognitive skills including analysis, synthesis, and professional judgment~\cite{cfa2025level3} over rote memorization. The exam's rigorous standards, reflected in typical pass rates of 55\%-65\%~\cite{cfa_passing_score_300hours}, make it an excellent benchmark for assessing advanced financial reasoning capabilities.


Previous studies~\cite{mahfouz2024state} showed that state-of-the-art models at the time of the analysis performed reasonably well on CFA Levels I and II mock exams but struggled significantly with Level III's advanced reasoning requirements, particularly on the essay questions. This gap reveals key limitations in the ability of the evaluated LLMs to handle sophisticated financial analysis—exactly the capabilities needed for real-world deployment in investment management.

However, prior evaluations have been limited in scope (typically 8-12 models), lack systematic prompting strategy comparison, and provide insufficient cost-effectiveness analysis for deployment decisions. Moreover, the rapid emergence of reasoning-focused models (o3-mini, DeepSeek-R1) and latest frontier models remains largely unevaluated on complex financial reasoning tasks.

This study addresses these critical gaps by providing the first comprehensive evaluation of 23 diverse LLMs on mock CFA Level III exams. Our evaluation spans frontier reasoning and non-reasoning models, both proprietary and open source, and uses a range of prompting strategies: zero-shot, chain-of-thought, and self-discover. Our key findings include:

\begin{itemize}
\item {\em First Demonstration of Professional-Grade LLM Performance:} We show that frontier models achieve composite scores of 79.1\% (o4-mini), and 75.9\% (Gemini 2.5 Pro) on mock CFA Level III exams, substantially exceeding the estimated 63\% passing threshold \cite{cfa_passing_score_300hours}—representing a significant milestone in financial Artificial Intelligence (AI) capabilities (see Table~\ref{tab:comprehensive_cfa_level3_performers}).

\item {\em Frontier models perform similarly on multiple choice questions but differ on essay questions.} When it comes to straightforward multiple choice questions, there is little to distinguish top models, but real differences emerge on essay questions, which require critical thinking and clear articulation of investment reasoning. Here leading reasoning models emerge as winners.

\item {\em Humans are more lenient than LLMs at grading essays.} Unlike the multiple choice questions, which can be graded objectively, essay questions require subjective assessment of the quality of the provided answer against  sample answers provided as part of the mock exams. When faced with such subjective evaluations, existing work~\cite{feuer2024style} has relied on a frontier LLM (e.g., GPT-4.1) as a judge, prompting it to provide a numeric grade according to a grading rubric. While scalable, this approach does not reflect real grading practices and suffers from potential LLM biases. We address this challenge by grading each answer to an essay question using both GPT-4 and a certified human grader. We find systematic differences between human and LLM grades, with the human graders awarding an average of 5.6 points more than the LLM-grader
on essay questions. Human and LLM-graders have perfect
agreement on about 70\% of the questions on average across
models, suggesting that the grade point differences arising
from around 30\% of the questions (see~Table~\ref{tab:all_models_comparison}). Despite these differences the overall finding of frontier LLMs passing the CFA Level III exam remain.

\item {\em How you ask determines what you get.} As observed in existing work \cite{he2024does}, 
the way the LLMs are prompted affects their performance significantly. We find that chain-of-thought prompting, where models explain their reasoning step-by-step, yielded the highest essay scores, while other approaches significantly underperformed (see Table~\ref{tab:combined_strategies}). This shows that effective AI financial advice requires sophisticated reasoning.

\item {\em Cost and speed matter for real-world deployment.} Most existing benchmarks consider objective task-specific performance measures of models, but real-world deployment must also consider cost and speed tradeoffs. Our analysis reveals significant accuracy-efficiency tradeoffs: advanced prompting improves MCQ accuracy by 7.8 percentage points but increases costs by 3-11x cost increases. Essay questions show similar cost and latency increases with advanced prompting. When deployed at scale in real world processing millions of user prompts each day, these differences in costs and latency easily add up. Our findings suggest an intermediate approach: deploying smaller, faster models for simpler queries but switching to frontier and more expensive models only for complex queries.

\end{itemize}

\section{Methodology}

\subsection{Dataset Construction and Validation}

As official CFA Level III questions post-2018 are not publicly available due to intellectual property protections, we constructed our dataset using mock exam materials--including questions, answers, and grading rubrics--from AnalystPrep, a reputable CFA preparation provider with over 100,000 candidates, following similar approaches in recent financial AI research~\cite{mahfouz2024state}. This paywall protection significantly reduces the chances of contamination of model training data with our test data. Our dataset comprises two components reflecting the actual exam structure:

\textbf{Multiple-Choice Questions (MCQs)} dataset, consisting of 60 questions, organized into 10 vignettes with 6 questions each, covering all major Level III curriculum areas.

\textbf{Essay Questions} dataset, comprising 11 unique vignettes with 43 total questions (149 total points) covering major Level III curriculum areas: Private Wealth Management (2 vignettes), Portfolio Management (2 vignettes), Private Markets (2 vignettes), Asset Management, Derivatives and Risk Management, Performance Measurement, Portfolio Construction, and Ethical and Professional Standards. Each vignette presents realistic financial scenarios followed by 2-5 open-ended questions requiring synthesis of multiple concepts and clear articulation of investment reasoning.

Data preprocessing involved Optical Character Recognition(OCR) extraction from PDF sources, conversion to structured JavaScript Object Notation(JSON) format, and expert review to confirm alignment with the official CFA curriculum standards and appropriate difficulty levels.

\subsection{Model Selection and Categorization}

Our benchmark includes 23 state-of-the-art LLMs, selected to represent a diverse range of capabilities, architectural designs, and provider ecosystems. These models are categorized as follows:

\textbf{Frontier Models:} This category includes highly capable, general-purpose models, often representing the cutting edge from major providers: o4-mini, o3-mini, GPT-4o, GPT-4.1 series (including base, mini, and nano variants), Grok 3 (including mini-high and mini-low )) , Claude-3.5-Sonnet, Claude-3.5-Haiku, Claude-3.7-Sonnet, Claude-Opus-4, Claude-Sonnet-4, Gemini 2.5 Pro, Gemini 2.5 Flash, and Mistral Large Official.

\textbf{Non-Reasoning Models:} These models do not include specific reasoning enhancements and are run with standard inference settings: GPT-4o, GPT-4.1, GPT-4.1-mini, GPT-4.1-nano, Claude-3.5-Sonnet, Claude-3.5-Haiku
Grok-3, Llama-3.1-8B-instant, Llama-3.3-70B, Llama-4-Maverick, Llama-4-Scout
Mistral-Large, Palmyra-Fin

\textbf{Reasoning-Enhanced Models:} These models are specifically designed or configured to improve reasoning capabilities by using their thinking feature: o3-mini, o4-mini, Deepseek-R1, Grok-3-mini-beta (evaluated with both high and low reasoning effort), Gemini 2.5 Pro, Gemini 2.5 Flash, Claude Opus 4, Claude Sonnet 4, and Claude 3.7 Sonnet.



\textbf{Open-Source Models:} This group comprises prominent open-source LLMs that offer accessible and adaptable alternatives. Models tested: Llama-3.1-8B instant, Llama-3.3-70B, Llama-4-Maverick, Llama-4-Scout and Deepseek-R1.

\textbf{Specialized Models:} We also tested the Palmyra-fin model, which has been trained and fine-tuned for the finance domain. 

\subsection{Prompting Strategy Design}

We evaluated three carefully designed prompting approaches to assess their impact on financial problem-solving performance on both MCQs and essays detailed in Appendix~\ref{subsec:template_variables}.

\textbf{Zero-Shot:} This is the most direct prompting strategy, where the model is given the relevant context and question and asked to provide a direct answer—selecting an option for MCQs or generating text for essays (see Appendix~\ref{subsec:mcq_zero_shot} and~\ref{subsec:essay_zero_shot} for the templates)—without explicit reasoning structures.

\textbf{Chain-of-Thought (CoT) with Self-Consistency (SC):} In this prompting strategy, we ask explicitly instruct the models to output a step-by-step reasoning process before providing the final answer (for MCQs, this involves selecting an option; for essays, generating the full text). To further increase the accuracy of responses, we employed {\em self-consistency}~\cite{wang2022self} by asking the model to generate N=3 or N=5 distinct reasoning paths and corresponding answers. 

For multiple-choice questions, we generated N diverse reasoning traces (N=3,5) for each question using model-specific temperature settings optimized per model configuration to ensure reasoning path diversity while maintaining logical coherence. Each sample followed Chain-of-Thought prompting, the template for which is detailed in Appendix~\ref{sec:appendix_mcq_prompts} with stochastic sampling parameters configured per model to generate varied reasoning paths while maintaining coherent output. We applied majority voting across the N samples to determine the final answer; in cases where vote counts were tied, we choose the first encountered option among tied candidates. 

  For essay questions, we adapted self-consistency to accommodate open-ended generation, using the prompt templates shown in Appendix~\ref{sec:appendix_essay_prompts} where traditional majority voting is infeasible. We generated N diverse essay responses (N=3,5) using temperature settings optimized per model based on their respective documentation to ensure 
  reasoning path diversity while maintaining coherent output. Each model then self-evaluated its own 
  response on a 0-10 scale using the prompt template shown in Appendix~\ref{subsec:grading_self}, providing confidence scores that serve as quality proxies for response 
  selection. We selected the essay with the highest self-confidence score as the final answer; in cases of tied scores, we selected the first response among the tied candidates to
  maintain deterministic selection. This approach maintains self-consistency's core principle of aggregating across multiple reasoning attempts while adapting the selection mechanism from discrete answer voting to confidence-based ranking. All selected essays were subsequently evaluated by an external GPT-4.1 judge using CFA Level III grading criteria to ensure unbiased assessment independent of the generating model's 
  self-evaluation.
  This prompting strategy more closely mimics how a human test taker might behave during the exam, going through a reasoning process before arriving at the final response. Simulating multiple reasoning paths effectively deals with inherent model stochasticity, providing a more robust assessment of model capabilities.


  \textbf{Self-Discover:} A metacognitive prompting strategy that guides models to explicitly construct their own reasoning methodology before problem-solving \cite{zhou2024self}. Unlike direct prompting, Self-Discover operates through explicit reasoning architecture construction: (1) Module Identification - the model first identifies which types of reasoning are required for the specific problem (e.g., causal analysis, quantitative calculation, comparative evaluation); (2) Module Contextualization - each identified reasoning type is then adapted to the specific problem context, with the model defining how it will apply each approach; (3) Structured Execution - the model systematically implements its self-designed reasoning framework; and (4) Critical Evaluation - final assessment using the established reasoning structure. This metacognitive approach differs from conventional prompting by requiring models to explicitly design their problem-solving methodology rather than directly attempting solutions.

For multiple-choice questions, Self-Discover adapts the four-stage metacognitive framework (see Appendix~\ref{subsec:mcq_self_discover} for the
  specific prompt) to constraint-bounded option evaluation. The strategy requires models to explicitly identify relevant reasoning modules—Definition
  Understanding, Quantitative Calculation, Conceptual Analysis, Comparative Evaluation, and Elimination Logic—before systematically implementing their
  self-constructed reasoning framework. This tests whether explicit reasoning architecture development enhances precision in scenarios where multiple
  options exhibit surface-level plausibility, requiring deep analytical discrimination rather than superficial pattern matching. Standardized answer
  formatting enables reliable automated evaluation while preserving the metacognitive rigor distinguishing Self-Discover from conventional prompting
  approaches.

    For essay questions, Self-Discover maintains the same four-stage metacognitive framework while 
  adapting the reasoning modules to accommodate comprehensive analytical writing tasks (the full 
  prompt structure is in Appendix~\ref{subsec:essay_self_discover}). This adaptation requires 
  models to explicitly structure essay-specific reasoning components—including problem 
  deconstruction, vignette analysis, and synthesis—before execution. This approach tests whether 
  explicit metacognitive essay planning yields superior analytical depth compared to implicit 
  reasoning strategies, requiring models to consciously architect their comprehensive 
  problem-solving methodology before implementation.



\subsection{Evaluation Metrics and Framework}

To evaluate model performance, we assigned a numerical grade to each model response and then aggregated them to obtain an overall numerical grade. MCQs were straightforward to grade: model generated answers were compared against the correct answers from the answer key to get the percentage of correct answers as the overall numerical grade.


Essay evaluation is more challenging because the free text responses of the models must be compared to the suggested response from the answer key on various qualitative factors, including semantic similarity, reasoning quality, and completeness. We address this challenge through three
   complementary evaluation approaches: (a) model self-grading, (b) automated external grading, and
   (c) human-expert grading.

\paragraph{(a) Model self-grading.}  
We implemented mechanisms for models to evaluate their own generated responses. In our adaptation of self-consistency to essay tasks, each model assigned confidence scores (0–10 scale) to its \(N\) generated responses, and the highest self-rated response was selected as the final answer. This self-evaluation serves two purposes: (1) enabling confidence-based response selection in our self-consistency strategy, and (2) providing insights into calibration, i.e., the alignment between self-assessed confidence and objective performance metrics.

\paragraph{(b) Automated external grading.}  
For automated external grading, we employed the LLM-as-a-judge framework \cite{zheng2023judging}, in which a frontier model assigns a numerical score to a response relative to a grading rubric and reference solution. LLM-judges are widely used for benchmarking because they provide scalable and consistent evaluations. Specifically, we used GPT-4.1 (temperature 0.0) as our primary LLM-judge for essay grading across all models and prompting strategies. Each grading prompt included the model output (with full reasoning traces where applicable), the reference answer, the question and vignette, the rubric, and the score range. The prompt enforced strict criteria-based scoring, disallowed subjective partial credit, and required integer-only scores (see Appendix~\ref{sec:appendix_grading_prompts}).  
We computed an overall essay score by taking a weighted average of individual essay scores, weighting by question point value, and normalizing to a 0–100 scale for comparability with MCQs.  
In addition, we report results from two standard automated metrics: cosine similarity, which measures vector-space similarity between model and reference answers, and ROUGE-L \cite{lin2004rouge}, which measures lexical overlap. Including these complementary metrics increases the robustness of our evaluation.

\paragraph{(c) Human-expert grading.}  
Although LLM-judges enable scalable evaluation, they have been shown to exhibit systematic biases, often prioritizing stylistic fluency over substantive correctness~\cite{feuer2024style}. To mitigate these limitations, we complement our automated grading strategies with
human-expert grading where we asked certified experts with
a great deal of experience real-world CFA Level III exams.
We report aggregated essay grades for both LLM-judges and
human-expert graders for all the models.

To implement human-expert grading, we focused on the self-consistency Chain-of-Thought (CoT) strategy with \(N=3\), both to control costs and because CoT (\(N=3\)) achieved the highest average performance across automated metrics for top-performing models. We assigned all 43 essay questions across 23 models—totaling \(43 \times 23 = 989\) responses—to a certified CFA Level III grader. Section~\ref{sec:human_llm_comparison} provides a detailed comparative analysis of human and LLM-assigned grades on a per-model basis.

\textbf{Overall Score Calculation:} Once we obtain aggregated scores for the MCQ and essay portions, we compute an overall score for the entire Level III mock test by taking a simple average of the two scores, reflecting equal weightage to both portions. This is consistent with how the points are allocated in the exam and aligned with previous work~\cite{mahfouz2024state}.




\begin{table*}[!htbp] 
\centering 
\small 
\setlength{\tabcolsep}{1pt} 
\begin{tabular}{@{}llccccccc@{}} 
\toprule 
Provider & Model & MCQ Score (Strat.) & Essay Score (Strat.) & Human Essay Score & Cosine Sim. & ROUGE-L & Overall Score \\ 
\midrule 
\multirow{2}{*}{OpenAI} & \cellcolor{gray!20}o4-mini & \cellcolor{gray!20}75.00 (Self-Discover) & \cellcolor{gray!20}\textbf{83.22} (CoT-SC N=3) & \cellcolor{gray!20}79.9 & \cellcolor{gray!20}0.4939 & \cellcolor{gray!20}\underline{0.1491} & \cellcolor{gray!20}\textbf{79.1} \\ 
 & \cellcolor{gray!20}o3-mini & \cellcolor{gray!20}75.00 (CoT-SC N=5) & \cellcolor{gray!20}73.83 (Self-Discover) & \cellcolor{gray!20}- & \cellcolor{gray!20}0.5532 &\cellcolor{gray!20} 0.0881 &\cellcolor{gray!20} 74.4 \\ 
 & \cellcolor{gray!20}o3-mini & \cellcolor{gray!20}75.00 (CoT-SC N=5) & \cellcolor{gray!20}73.83 (CoT-SC N=5) & \cellcolor{gray!20}- & \cellcolor{gray!20}0.5493 & \cellcolor{gray!20}0.1179 & \cellcolor{gray!20}74.4 \\ 
\cmidrule(lr){1-8} 
\multirow{2}{*}{Google} & \cellcolor{gray!20}Gemini 2.5 Flash & \cellcolor{gray!20}73.33 (Zero Shot) & \cellcolor{gray!20}\underline{81.21} (CoT-SC N=3) & \cellcolor{gray!20}78.5 & \cellcolor{gray!20}\textbf{0.5793} & \cellcolor{gray!20}0.0847 & \cellcolor{gray!20}\underline{77.3} \\ 
 & \cellcolor{gray!20}Gemini 2.5 Pro & \cellcolor{gray!20}\textbf{76.67} (Zero Shot) & \cellcolor{gray!20}75.17 (CoT-SC N=3) & \cellcolor{gray!20}83.2 & \cellcolor{gray!20}\underline{0.5695} & \cellcolor{gray!20}0.0774 & \cellcolor{gray!20}75.9 \\ 
 & \cellcolor{gray!20}Gemini 2.5 Pro & \cellcolor{gray!20}75.00 (CoT-SC N=3) & \cellcolor{gray!20}75.17 (CoT-SC N=3) & \cellcolor{gray!20}83.2 & \cellcolor{gray!20}0.5695 & \cellcolor{gray!20}0.0774 & \cellcolor{gray!20}75.1 \\ 
 & \cellcolor{gray!20}Gemini 2.5 Pro & \cellcolor{gray!20}76.67 (Self-Discover) & \cellcolor{gray!20}61.74 (Self-Discover) & \cellcolor{gray!20} - & \cellcolor{gray!20}0.5691 & \cellcolor{gray!20}0.0704 & \cellcolor{gray!20}69.2 \\ 
\cmidrule(lr){1-8} 
\multirow{2}{*}{Anthropic} &\cellcolor{gray!20} Claude Opus 4 & \cellcolor{gray!20}73.33 (CoT-SC N=3) & \cellcolor{gray!20}76.51 (CoT-SC N=3) & \cellcolor{gray!20}81.9 & \cellcolor{gray!20}0.5465 & \cellcolor{gray!20}0.1256 & \cellcolor{gray!20}74.9 \\ 
 & \cellcolor{gray!20}Claude Sonnet 4 & \cellcolor{gray!20}75.00 (CoT-SC N=3) & \cellcolor{gray!20}71.81 (CoT-SC N=3) & \cellcolor{gray!20}73.8 & \cellcolor{gray!20}0.5238 & \cellcolor{gray!20}0.1241 & \cellcolor{gray!20}73.4 \\ 
 & \cellcolor{gray!20}Claude 3.7 Sonnet & \cellcolor{gray!20}73.33 (Self-Discover) & \cellcolor{gray!20}67.79 (CoT-SC N=3) & \cellcolor{gray!20}77.8 & \cellcolor{gray!20}0.5410 & \cellcolor{gray!20}0.1172 & \cellcolor{gray!20}70.6 \\ 
\cmidrule(lr){1-8} 
\multirow{2}{*}{xAI} & Grok-3 & 71.67 (CoT-SC N=3) & 75.84 (CoT-SC N=5) &  - & 0.5648 & 0.0945 & 73.8 \\ 
& \cellcolor{gray!20}Grok-3 Mini Low Effort & \cellcolor{gray!20}75.00 (CoT-SC N=3) & \cellcolor{gray!20}69.80 (CoT-SC N=3) & \cellcolor{gray!20}71.8 & \cellcolor{gray!20}0.5403 & \cellcolor{gray!20}0.0738 & \cellcolor{gray!20}72.4 \\ 
\cmidrule(lr){1-8} 
\multirow{1}{*}{DeepSeek} & \cellcolor{gray!20}DeepSeek-R1 & \cellcolor{gray!20}71.67 (Self-Discover) & \cellcolor{gray!20}63.09 (CoT-SC N=3) & \cellcolor{gray!20}62.4 & \cellcolor{gray!20}0.5383 & \cellcolor{gray!20}0.1132 & \cellcolor{gray!20}67.4 \\ 
\cmidrule(lr){1-8} 
\multirow{1}{*}{Writer} & Palmyra-fin & 68.33 (Self-Discover) & 58.39 (Zero Shot) &  - & 0.5376 & \textbf{0.1652} & 63.4 \\ 
\bottomrule 
\end{tabular} 
\captionsetup{width=\textwidth} 
\caption{CFA Level III top model performance summary (subset of 23 evaluated models shown for space). MCQ and Essay scores reflect best performance across all strategies tested (strategy noted in parentheses), with all scores reported as percentages. Models may appear multiple times when different strategy combinations yield distinct overall scores. Human Essay scores (graded by a CFA expert) reflect evaluation of responses generated using the best-performing essay strategy: Self-Consistency with 3 samples (CoT-SC N=3); human evaluation was limited to this approach. Cosine Similarity and ROUGE-L scores correspond to the essay strategy associated with the listed Essay Score. Overall Score is the average of MCQ Score and normalized Essay Score. Bold indicates best performance, underlined indicates second-best. Gray highlighting denotes non-reasoning models.}
\label{tab:comprehensive_cfa_level3_performers} 
\end{table*}

\textbf{Efficiency and Cost Metrics:} In addition to model capability metrics, we also tracked key operational and economic metrics for all the models: average API latency, input/output token usage (including internal reasoning tokens where available), estimated cost per evaluation (USD), average essay answer length, and total run time per model-strategy pair. These efficiency and cost metrics allow us to understand the trade off that exists in model capabilities and costs, which are critical for large-scale practical implementation of these models.

\section{Results}

\begin{figure}[htbp]
\centering
\includegraphics[width=0.5\textwidth]{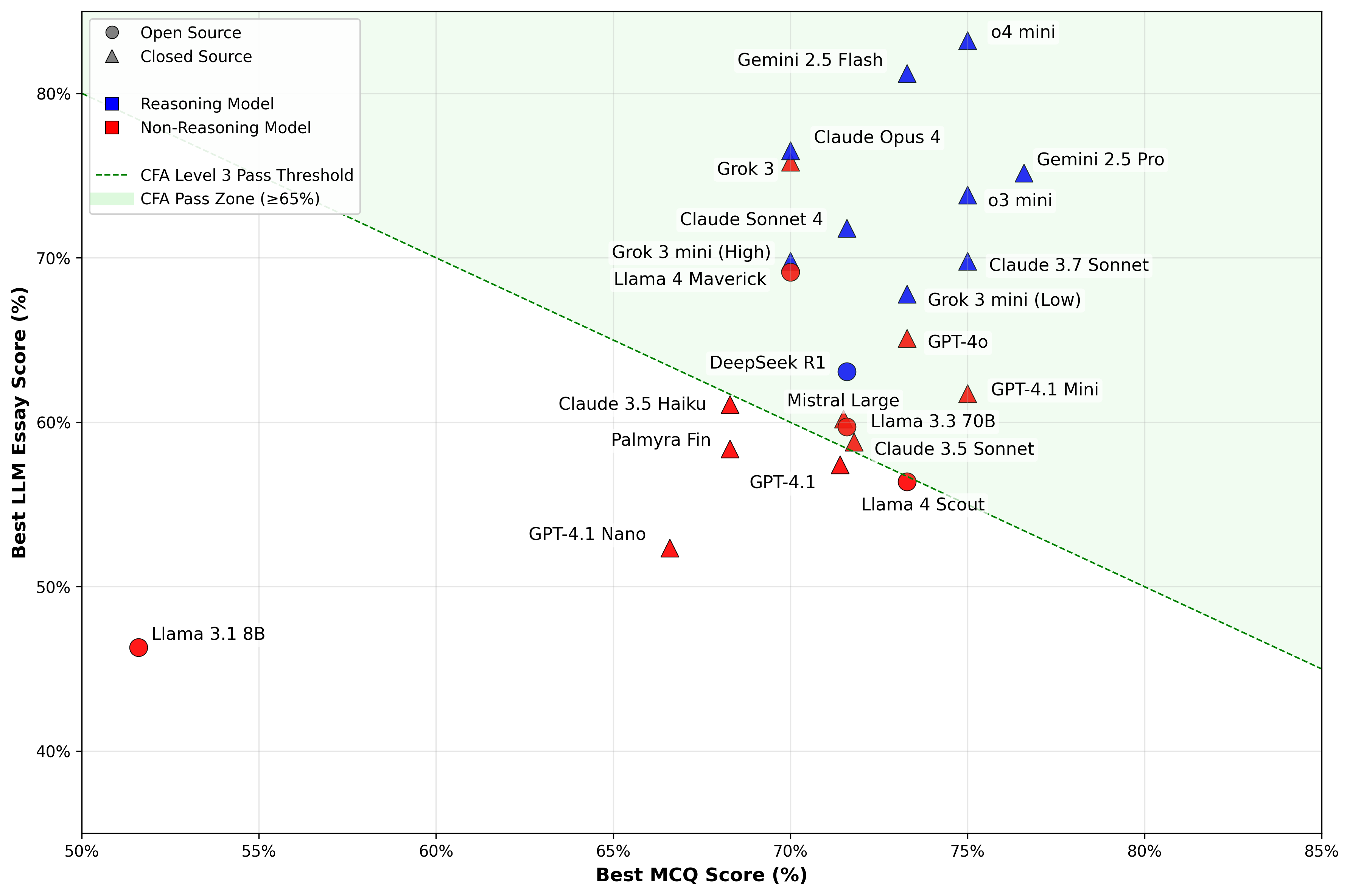}
\caption{MCQ vs. Essay performance correlation across all evaluated models, plotting best MCQ accuracy (x-axis) against best Essay Score (y-axis). Key: Blue: Reasoning, Red: Non-reasoning; Triangle: Closed-source, Circle: Open-source. The green hue indicates models achieving the Minimum Passing Score (MPS) of $>$=65\% for the CFA Level III examination.}
\label{fig:mcq_llm_essay_correlation}
\end{figure}

\begin{figure}[htbp]
\centering
\includegraphics[width=0.5\textwidth]{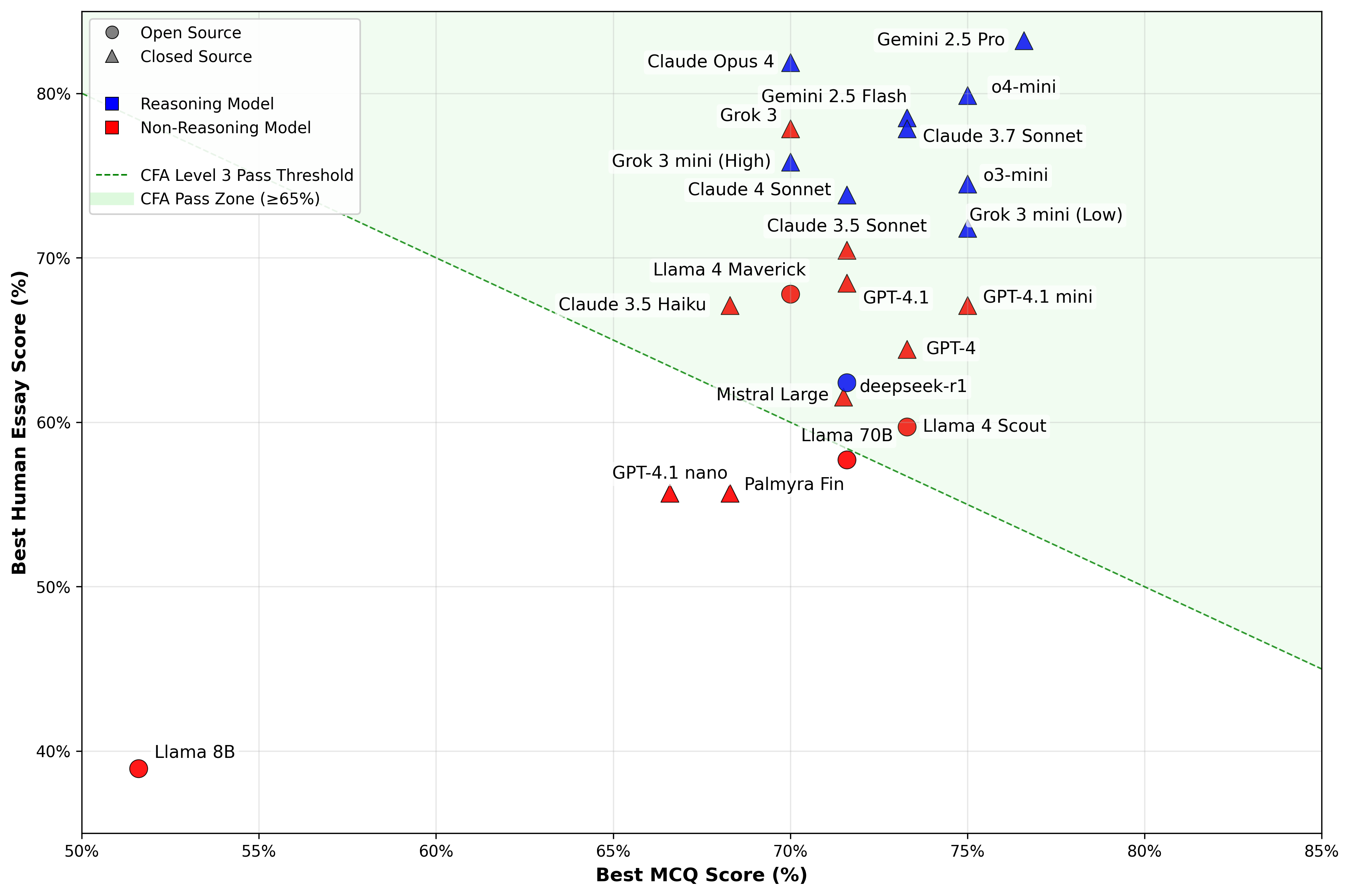}
\caption{MCQ vs. Essay performance correlation across all evaluated models, plotting best MCQ accuracy (x-axis) against Human Graded CoT with Self Consistency (\(N=3\)) Essay Scores (y-axis). Key: Blue: Reasoning, Red: Non-reasoning; Triangle: Closed-source, Circle: Open-source. The green hue indicates models achieving the Minimum Passing Score (MPS) of $>$=65\% for the CFA Level III examination.}
\label{fig:mcq_human_essay_correlation}
\end{figure}

We now discuss our findings. As mentioned above, we tested 23 leading LLMs in three different prompting strategies, resulting in a total of $23 \times 3 = 69$ model-prompt combinations. Figures~\ref{fig:mcq_llm_essay_correlation} and \ref{fig:mcq_human_essay_correlation} and Table~\ref{tab:comprehensive_cfa_level3_performers} summarize the MCQ and essay performances of the top performing subset of the 69 model-prompt combinations. Figure~\ref{fig:mcq_llm_essay_correlation} shows a scatterplot of MCQ and LLM-judged essay scores whereas Figure~\ref{fig:mcq_human_essay_correlation} shows a similar scatterplot of the MCQ scores against human-expert essay grades. Both figures highlight in green the region above the minimum passing standard (MPS),indicating model performance at or above the level required to pass the CFA Level III exam. We adopt an MPS of 65\%~\cite{cfa_passing_score_300hours}. A model is considered to have passed the CFA Level III exam if its overall score—defined as the average of its multiple-choice and essay scores—is at least 65\%.
Table~\ref{tab:comprehensive_cfa_level3_performers} provides more details, including the cosine similarity and ROUGE-L scores.

Our results clearly demonstrate that a host of leading prompt-model combinations comfortably pass mock CFA Level III exams. Visually it is clear that there is a predominance of closed-source (shown as triangles) and reasoning (colored in blue) models among the best performing ones. From examining Table~\ref{tab:comprehensive_cfa_level3_performers}, it is clear that the top performers all score around 75\% on MCQs and their scores seem to be clustered in the range of 71-75\%. However, these models exhibit quite a bit of variance on their essay scores. Previous work~\cite{mahfouz2024state} has found that the state-of-the-art (SOTA) models at the time of the analysis performed well on MCQs but struggled on the essay questions. Our analysis reveals that while top models achieve convergent performance on MCQs, essay questions exhibit greater variance and thus provide better discrimination between model capabilities.



While most existing works adopt LLM-judges to evaluate LLM performance, we use both LLM-judged and human-expert grades. We find that for the same answers, LLM-judges and human-experts often assign different numerical grades. However, our core finding that a host of frontier models clear mock CFA-Level III exam remains true across both grading strategies. In Section~\ref{sec:human_llm_comparison}, we carry out a more detailed comparison between how human-expert grades differ from those of LLM-judges.

\subsection{Model Type Analysis}

In Table~\ref{tab:model_type_comparison_summary}, we compare the overall performance of reasoning-enhanced models against non-reasoning models. We find that reasoning-enhanced models consistently outperform non-reasoning models across all performance metrics. For MCQ tasks, reasoning models achieved 73.1\% accuracy compared to 69.4\% for non-reasoning models. The performance gap is even more pronounced on essay questions: on LLM-judged grades, reasoning-enhanced models scored 73.7\% compared to 61.8\% for non-reasoning models, representing a 19.1\% improvement. Human evaluators showed similar patterns, rating reasoning-enhanced model outputs at 75.97\% compared to 62.52\% for non-reasoning models. This finding is consistent with existing work showcasing that reasoning capabilities incur significant computational times (almost 3x for essay questions) but deliver measurable performance improvements on challenging analytical tasks.

\begin{table}[H]
\centering
\small
\setlength{\tabcolsep}{5pt}
\begin{tabular}{@{}lccccc@{}}
\toprule
\multirow{2}{*}{Model Type} & \multicolumn{2}{c}{MCQ Performance} & \multicolumn{3}{c}{Essay Performance} \\
\cmidrule(lr){2-3} \cmidrule(lr){4-6}
 & Accuracy & Time(s) & LLM & Human & Time(s) \\
\midrule
Reasoning & \textbf{0.731} & 59.2 & \textbf{73.71} & \textbf{75.97} & 102.8 \\
Non-Reasoning & 0.694 & \textbf{17.9} & 61.87 & 62.52 & \textbf{31.8} \\
\bottomrule
\end{tabular}
\caption{Performance comparison by architecture type across MCQ and Essay tasks (avg. values for Self consistency n = 3). MCQ accuracy is on a 0-1 scale, Human Score and Essay Score on a 0-100 scale. Human Scores are on Self-Consistency CoT n=3. Time shows average processing seconds. Bold indicates superior performance.}
\label{tab:model_type_comparison_summary}
\end{table}

\subsection{Prompting Strategy Effectiveness and Efficiency Trade-offs}

Advanced prompting strategies show mixed but generally positive results, with substantial implications for cost-effectiveness in real-world deployment. Table~\ref{tab:combined_strategies} shows that Chain-of-Thought with Self-Consistency (CoT-SC \(N=5\)) achieves the highest MCQ accuracy at 69.6\%, representing a 7.8 percentage point improvement over Zero Shot prompting (61.8\%). However, this performance gain comes at significant computational cost—requiring 10.5x longer processing time (63.1s vs 6.0s) and 11.1x higher cost (\$2.84 vs \$0.255) compared to Zero Shot strategies (see Table~\ref{tab:combined_efficiency_strategies}). 

For essay questions, the pattern is more nuanced. CoT-SC (\(N=5\)) achieves the highest mean Essay Score (61.03\%) and strongest semantic alignment (0.5376 cosine similarity), but requires 8.6x longer processing times (130.48s vs 15.18s) and 8.8x higher evaluation costs (\$3.390 vs \$0.387) compared to Zero-shot approaches. Interestingly, Zero Shot prompting yields the highest ROUGE-L F1 score (0.1522), suggesting closer lexical overlap with reference answers despite lower semantic similarity scores.

Self-Discover prompting presents a compelling middle ground for MCQ questions, delivering a substantial 6.8 percentage point accuracy improvement over Zero Shot while increasing costs by only 2.7x. However, for essay evaluation, Self-Discover significantly underperforms across all metrics, suggesting that its meta-cognitive planning approach may not align well with the precise, rubric-adherent answers required for the CFA Level III exam.


\begin{table}
\centering
\footnotesize
\setlength{\tabcolsep}{3pt}
\begin{tabular}{@{}lccccccc@{}}
\toprule
\multirow{2}{*}{Strategy} & \multicolumn{3}{c}{MCQ Performance} & \multicolumn{3}{c}{Essay Performance} \\
\cmidrule(lr){2-4} \cmidrule(lr){5-7}
 & Acc. & Time & Cost & Score & Cos.Sim. & ROUGE-L \\
\midrule
Zero Shot & 61.8 & 6.0 & 0.255 & 57.35 & 0.5151 & \textbf{0.1522} \\
Self-Discover & 68.6 & 14.9 & 0.686 & 46.32 & 0.5012 & 0.0977 \\
CoT-SC (N=3) & 69.1 & 37.9 & 1.701 & \underline{60.40} & \underline{0.5353} & 0.1227 \\
CoT-SC (N=5) & \textbf{69.6} & 63.1 & 2.840 & \textbf{61.03} & \textbf{0.5376} & \underline{0.1228} \\
\bottomrule
\end{tabular}
\caption{Combined MCQ and essay prompting strategy comparison. MCQ metrics: Accuracy (\%), processing time (seconds), and cost (USD). Essay metrics: LLM-assessed score (0-100 scale), cosine similarity (0-1 scale), and ROUGE-L F1 (0-1 scale). All values represent means aggregated across evaluated models. Bold indicates best performance, underlined shows second best.}
\label{tab:combined_strategies}
\end{table}

\begin{table}
\centering
\footnotesize
\setlength{\tabcolsep}{6pt}
\begin{tabular}{@{}lcccc@{}}
\toprule
\multirow{2}{*}{Strategy} & \multicolumn{2}{c}{MCQ Performance} & \multicolumn{2}{c}{Essay Performance} \\
\cmidrule(lr){2-3} \cmidrule(lr){4-5}
 & Latency (s) & Cost (\$) & Latency (s) & Cost (\$) \\
\midrule
Zero Shot & 6.0 & 0.255 & 15.18 & 0.387 \\
Self-Discover & 14.9 & 0.686 & 25.27 & 0.703 \\
CoT-SC (N=3) & 37.9 & 1.701 & 69.83 & 2.089 \\
CoT-SC (N=5) & 63.1 & 2.840 & 130.48 & 3.390 \\
\bottomrule
\end{tabular}
\caption{Comprehensive efficiency comparison across MCQ and essay tasks. MCQ metrics and Essay metrics show average latency and cost per task. Values aggregated across all 23 models for MCQ and all models for essays.}
\label{tab:combined_efficiency_strategies}
\end{table}

\subsection{Open-Source versus Closed-Source Model Comparison}
\label{sec:opensource_vs_closedsource_comparison}



The aggregated analysis reveals that closed-source models achieve marginally superior performance compared to open-source alternatives. Closed-source models demonstrated higher average MCQ accuracy (72.69\% versus 70.00\%) and LLM-graded essay quality (67.80\% versus 58.93\%). Conversely, open-source models exhibited slight advantages in lexical similarity metrics, achieving marginally higher mean Cosine Similarity (0.5413 versus 0.5405) and ROUGE-L F1 scores (0.1512 versus 0.1499). These findings highlight the performance-accessibility trade-off inherent in model selection, with proprietary models offering superior analytical capabilities while open-source alternatives provide competitive semantic alignment at potentially lower deployment costs.

Our comprehensive analysis demonstrates that effective deployment of LLMs for CFA-level financial reasoning requires careful consideration of the accuracy-efficiency trade-off. While advanced prompting strategies and reasoning-enhanced models deliver measurable performance improvements, the substantial increases in computational cost and processing time suggest that a tiered approach may be optimal: deploying simpler, faster models for routine queries while reserving advanced reasoning capabilities for complex analytical tasks.

\section{Alignment between human and LLM Grades}
\label{sec:human_llm_comparison}

In this section, we analyze the degree of alignment between essay grades assigned by a certified CFA human-grader and the LLM-grader. To keep costs manageable, we focused on one of the best performing prompting strategies, Chain-of-Thought with Self-Consistency with three iterations, CoT (N=3). Both human and LLM graders graded all of the 43 essay responses generated by each of the 23 models, resulting in grades for a total of 989 question answer pairs (43 questions $\times$ 23 models).

Our design ensures availability of both human and LLM assigned grades for the {\em same} set of responses with 100\% coverage across all questions, eliminating possible sampling biases. It also allows for response-by-response comparisons. We measure alignment using the following metrics, reported per model:


\begin{itemize}
    \item \textbf{Alignment Gap ($\Delta$)}: Human - LLM grade difference
    \item \textbf{Variance}: Variance of grade differences across questions
    \item \textbf{Agreement Rate}: \% of exact grade matches across questions
\end{itemize}
Note that our alignment gap measure subtracts LLM grades from human grades, so the sign indicates the relative leniency of each of the graders.

\begin{table}[h]
\centering
\setlength{\tabcolsep}{1pt}
\begin{tabular}{lcccccc}
\toprule
Model & Human & LLM & $\Delta$ & Var. & Agree \% \\
\midrule
\cellcolor{gray!20}Gemini 2.5 pro       & \cellcolor{gray!20}83.2 & \cellcolor{gray!20}75.2 & \cellcolor{gray!20}+8.0  & \cellcolor{gray!20}1.635 & \cellcolor{gray!20}69.8 \\
\cellcolor{gray!20}Claude Opus 4        & \cellcolor{gray!20}81.9 & \cellcolor{gray!20}76.5 & \cellcolor{gray!20}+5.4  & \cellcolor{gray!20}1.441 & \cellcolor{gray!20}76.7 \\
\cellcolor{gray!20}o4 mini              & \cellcolor{gray!20}79.9 & \cellcolor{gray!20}83.2 & \cellcolor{gray!20}-3.3  & \cellcolor{gray!20}2.010 & \cellcolor{gray!20}76.7 \\
\cellcolor{gray!20}Gemini 2.5 Flash     &\cellcolor{gray!20} 78.5 & \cellcolor{gray!20}81.2 &\cellcolor{gray!20} -2.7  &\cellcolor{gray!20} 1.277 &\cellcolor{gray!20} 72.1 \\
Grok 3               & 77.9 & 73.2 & +4.7  & 1.520 & 79.1 \\
\cellcolor{gray!20}Claude 3.7 Sonnet    & \cellcolor{gray!20}77.9 & \cellcolor{gray!20}67.8 & \cellcolor{gray!20}+10.1 & \cellcolor{gray!20}1.423 & \cellcolor{gray!20}81.4 \\
\cellcolor{gray!20}Grok 3 mini high effort & \cellcolor{gray!20}75.8 & \cellcolor{gray!20}69.8 & \cellcolor{gray!20}+6.0  & \cellcolor{gray!20}1.503 & \cellcolor{gray!20}72.1 \\
\cellcolor{gray!20}o3 mini              & \cellcolor{gray!20}76.0 & \cellcolor{gray!20}72.5 & \cellcolor{gray!20}+3.5  & \cellcolor{gray!20}1.828 & \cellcolor{gray!20}67.4 \\
\cellcolor{gray!20}Grok 3 mini low effort & \cellcolor{gray!20}75.2 & \cellcolor{gray!20}72.7 & \cellcolor{gray!20}+2.5  & \cellcolor{gray!20}0.733 & \cellcolor{gray!20}74.4 \\
\cellcolor{gray!20}Claude sonnet 4      & \cellcolor{gray!20}73.8 & \cellcolor{gray!20}71.8 & \cellcolor{gray!20}+2.0  & \cellcolor{gray!20}0.543 & \cellcolor{gray!20}79.1 \\
Claude 3.5 sonnet    & 70.5 & 57.7 & +12.8 & 1.443 & 76.7 \\
Claude 3.5 haiku     & 67.1 & 61.1 & +6.0  & 1.169 & 69.8 \\
GPT 4.1 mini         & 67.1 & 60.4 & +6.7  & 0.992 & 69.8 \\
Llama 4 Maverick*    & 67.8 & 57.7 & +10.1 & 1.756 & 65.1 \\
GPT 4.1              & 68.5 & 54.4 & +14.1 & 1.684 & 67.4 \\
GPT 4o               & 64.4 & 61.1 & +3.3  & 1.677 & 65.1 \\
Mistral Large        & 61.7 & 56.4 & +5.3  & 1.250 & 67.4 \\
Llama 4 Scout*       & 59.7 & 55.7 & +4.0  & 0.885 & 65.1 \\
\cellcolor{gray!20}DeepSeek r1*          & \cellcolor{gray!20}62.4 & \cellcolor{gray!20}63.1 & \cellcolor{gray!20}-0.7  & \cellcolor{gray!20}0.928 & \cellcolor{gray!20}72.1 \\
Llama 3.3 70b*       & 57.7 & 51.7 & +6.0  & 1.027 & 69.8 \\
GPT 4.1 nano         & 55.7 & 52.4 & +3.3  & 1.772 & 67.4 \\
Palmyra Fin          & 55.7 & 50.3 & +5.4  & 1.060 & 74.4 \\
Llama 3.1 8b*        & 38.9 & 31.5 & +7.4  & 1.100 & 62.8 \\
\midrule
Overall Average      & 68.4 & 62.8 & +5.6  & 1.344 & 70.4 \\
\bottomrule
\end{tabular}
\caption{Comparison of Human and LLM Grades for All Models (SC-CoT n=3) with Alignment Gap ($\Delta$), Variance , and Agreement Rate (\%). Reasoning models are highlighted in gray. Open source models are marked with an asterisk (*).}
\label{tab:all_models_comparison}
\end{table}

Table~\ref{tab:all_models_comparison} presents our results. The results reveal systematic patterns in human-LLM grading alignment, with human graders consistently assigning higher scores than the automated LLM grader across a majority of models, with the the LLM assigning more lenient . Specifically, human graders assign an average of 5.6 points more than the LLM-grader on essay questions. Human and LLM-graders have perfect agreement on about 70\% of the questions on average across models, suggesting that the grade point differences arising from around 30\% of the questions.
Notably, only three models demonstrate LLM-favored grading bias: o4 mini (-3.3 points), Gemini 2.5 Flash (-2.7 points), and DeepSeek r1 (-0.7 points). 

While our data does not support definitive conclusions on the reasons for such systematic differences, we can speculate two possible reasons. First, our grading prompt (see Appendix ~\ref{sec:appendix_grading_prompts}) asks the LLM to be a ``strict'' grader and ``follow the provided grading details EXACTLY.'' Second, the correct response on the answer key, included as a reference in the grading prompt, generally is more concise than the model-generated response. Both of these aspects in conjunction might lead the LLM-grader to be less lenient than a human-grader. 

Nevertheless, our finding highlights that LLM-based automated evaluation strategies are sensitive to the grading prompt and exhibit potential systematic biases.  While such automated evaluation strategies are commonly used in practice, one must carefully trade off the cost and scale benefits against potential biases introduced into the evaluation metrics.

\section{Conclusions and Future Work}

\subsection{Key Findings and Implications}

Our work is the first to comprehensively evaluate 23 LLMs on the CFA Level III exam. We find that frontier models now achieve professional-grade performance on specialized, high-stakes financial reasoning tasks, with leading models consistently exceeding the estimated 63\% passing threshold for CFA Level III. 

Our results showcase that while model capabilities cluster around 71-75\% for multiple-choice questions, they exhibit substantial variance on essay questions, which require complex reasoning and synthesis. This suggests that simpler, more straightforward tasks have become commoditized across models, whereas complex and nuanced reasoning tasks still differentiate frontier and reasoning-enhanced models from their peers.

While most existing work uses LLM-based automated evaluation strategies, we are the first to evaluate model performances using both LLM-based and human-expert based grading. We find systematic differences between LLM and human-expert graders, with LLM graders consistently assigning lower scores than human experts (5.6 points lower on average), likely because stricter prompt instructions and concise reference answers influence their assessments. This finding highlights how practitioners must carefully balance the cost and scale benefits of automated evaluation against potential systematic biases that LLM-based grading introduces.

Finally, the prompting strategy critically determines LLM performance, with chain-of-thought approaches resulting in substantial performance improvements at 3-11x cost increases. Such cost considerations must be carefully balanced against performance gains during practical deployment.

\subsection{Future Research}

Our comprehensive evaluation framework opens several high-priority research directions that could advance financial AI: 

{\bf Cross-Certification Validation:} Our work opens the doors to evaluate LLM performance on other financial certifications, such as FRM, CAIA, and CPA. This will test whether LLM capabilities generalize across diverse financial domains and provide broader insights into how useful they are for specialized professional assessments.

{\bf Enhanced Self-Assessment Frameworks:} We currently use simple confidence scoring (0-10 scale) to select responses in our self-consistency implementation. Future work could explore more sophisticated approaches where models provide detailed justifications, compare their responses against reference standards, and explicitly identify strengths and weaknesses in their answers. One could also implement JSON-formatted outputs that mirror expert assessment patterns to enable more reliable autonomous decision-making.

{\bf Comprehensive Human Expert Validation:} Due to cost considerations, we focused our human expert evaluation only on one prompting strategy, CoT-SC (N=3). Future studies could extend human expert evaluation to all prompting strategies, allowing us to understand strategy-specific evaluation biases that could inform more accurate automated assessment systems.


{\bf Prompt Engineering Optimization:} Future work should systematically investigate how formatting elements impact financial reasoning performance, including emphasis methods, structural organization, and visual hierarchy design. This research could reveal whether specific presentation choices significantly influence reasoning quality across different model architectures.

LLMs continue evolving rapidly, making rigorous domain-specific benchmarks like ours essential to track progress toward professional-grade financial AI capabilities. Our open-source evaluation framework~\cite{cfa_essay_reproducer_code,cfa_mcq_reproducer_code}  provides a foundation for continued research into making LLMs more capable and cost-effective tools for financial professionals, while maintaining the high standards necessary for responsible deployment in this critical domain.

\section*{Ethical Statement}

This research evaluates publicly available language models using educational materials, presenting no direct ethical concerns. However, our findings have important implications for responsible AI deployment in finance. The performance gaps we identified underscore the critical need for continued human oversight in financial decision-making, particularly given the fiduciary responsibilities involved in investment management. We emphasize that current LLMs should be viewed as assistive tools rather than autonomous decision-makers for financial applications.

\section*{Acknowledgments}

This research was supported in part by GoodFin and New York University (NYU) Faculty Research. We thank the CFA Institute for their educational materials and mission, AnalystPrep for access to mock examination content, and the broader financial education community for maintaining rigorous professional standards that enable meaningful AI evaluation.

\bibliographystyle{named}
\bibliography{ijcai25}

\onecolumn
\appendix
\section*{Appendix}

\appendix
\section{Template Variables}
  \label{subsec:template_variables}

  Our evaluation framework employs standardized template variables across all prompt configurations
   to ensure consistency and reproducibility.

  \subsubsection{Multiple-Choice Question (MCQ) Variables}
  Used in MCQ prompting strategies for discrete answer selection:
  \begin{itemize}
      \item \texttt{\{vignette\}}: The contextual scenario providing background information for
  analysis
      \item \texttt{\{question\_stem\}}: The primary question requiring a single letter response
  (A, B, or C)
      \item \texttt{\{option\_a\}}, \texttt{\{option\_b\}}, \texttt{\{option\_c\}}: The three
  possible answer choices
  \end{itemize}

  \subsubsection{Essay Question Variables}
  Used in essay prompting strategies for comprehensive written responses:
  \begin{itemize}
      \item \texttt{\{folder\}}: The CFA curriculum topic category (e.g., Portfolio Management,
  Ethics)
      \item \texttt{\{vignette\}}: The contextual scenario or case study providing analytical
  foundation
      \item \texttt{\{question\}}: The essay question requiring comprehensive analysis and
  structured response
  \end{itemize}

  \subsubsection{Grading Variables}
  Used for automated LLM-as-judge evaluation of essay responses:
  \begin{itemize}
      \item \texttt{\{question\}}: The original essay question for grading context
      \item \texttt{\{vignette\}}: The contextual scenario used by the grader for reference
      \item \texttt{\{answer\_grading\_details\}}: CFA Level III rubric specifying exact scoring
  criteria
      \item \texttt{\{correct\_answer\}}: Expert reference answer for grader guidance
      \item \texttt{\{generated\_answer\}}: The model-produced response being evaluated
      \item \texttt{\{min\_score\}}, \texttt{\{max\_score\}}: Score boundaries for the specific
  question
  \end{itemize}

\clearpage

\section{Multiple-Choice Question Prompt Templates}
\label{sec:appendix_mcq_prompts}

\subsection{Zero-Shot Prompt Template}
\label{subsec:mcq_zero_shot}

\begin{figure}[h!]
\centering
\fbox{\begin{minipage}{0.95\textwidth}
\small\ttfamily
\noindent Your task is to answer the following multiple-choice question.\\
Your response MUST be ONLY the single letter of the correct option (A, B, or C).\\
Do NOT include any other text, reasoning, formatting, or explanation.\\[0.5em]

Vignette:\\
\{vignette\}\\[0.5em]

Question Stem:\\
\{question\_stem\}\\[0.5em]

Options:\\
A: \{option\_a\}\\
B: \{option\_b\}\\
C: \{option\_c\}\\[0.5em]

Carefully read the vignette, question, and options. Choose the single best answer.\\[0.5em]

Answer (select one letter ONLY: A, B, or C):
\end{minipage}}
\caption{Zero-shot Prompt Template for multiple-choice questions.}
\label{fig:prompt_mcq_default}
\end{figure}

\subsection{Chain-of-Thought (CoT) Prompt Template}
\label{subsec:mcq_cot}

\begin{figure}[h!]
\centering
\fbox{\begin{minipage}{0.95\textwidth}
\small
\small\ttfamily

You are a Chartered Financial Analyst (CFA) charterholder. Your task is to answer one multiple‐choice question from the CFA curriculum. Follow these steps:\\

1. Restate the question stem in your own words.\\

2. Think through it step by step, showing your reasoning (use bullet points if helpful).\\

3. Evaluate each of the choices provided in the Options section, noting why each could be right or wrong.\\

4. Conclude by selecting the single best answer. First, provide a one-sentence justification for your choice. Then, on a new, separate line, write "Final Answer: [LETTER]", where [LETTER] is the capital letter of your chosen option (e.g., "Final Answer: A"). This "Final Answer: [LETTER]" line must be the absolute last line of your response.\\

Vignette:\\
\{vignette\}\\[0.5em]

Question Stem:\\
\{question\_stem\}\\[0.5em]

Options:\\
A: \{option\_a\}\\
B: \{option\_b\}\\
C: \{option\_c\}\\[0.5em]

Answer:

\end{minipage}}
\caption{Chain-of-Thought Prompt Template for multiple-choice questions.}
\label{fig:prompt_mcq_cot}
\end{figure}

\clearpage
\subsection{Self-Discover Prompt Template}
\label{subsec:mcq_self_discover}

\begin{figure}[h!]
\centering
\fbox{\begin{minipage}{0.95\textwidth}
\small
\small\ttfamily

**Task:** Solve the following multiple-choice question by first devising a reasoning structure using the Self-Discover method.\\

**Context/Vignette:**\\
\{vignette\}\\[0.5em]

**Question Stem:**\\
\{question\_stem\}\\[0.5em]

**Options:**\\
A: \{option\_a\}\\
B: \{option\_b\}\\
C: \{option\_c\}\\[0.5em]

**Self-Discover Reasoning Process:**\\

**1. Select Reasoning Modules:**
   Identify and list the core reasoning modules or types of thinking needed to solve this specific question. Examples: Causal Reasoning, Definition Understanding, Calculation, Comparison, Rule Application, Concept Identification, etc.\\

**2. Adapt Modules to the Problem:**
   For each selected module, briefly explain how it specifically applies to this question and the given options. Outline the steps you will take within each module.\\

**3. Implement Reasoning Structure:**
   Execute the plan outlined above step-by-step.
   [Model generates reasoning steps here]\\

**4. Final Answer:**
   Based on the reasoning, critically evaluate the options and provide the final answer.
   **IMPORTANT**: Conclude your response with the final answer choice letter (A, B, or C) on a new line, formatted exactly as: `The final answer is: **[Option Letter]**` (e.g., `The final answer is: **B**`). Do not include any other text after this final line.
   [Model provides final answer letter here in the specified format]\\

**Begin Reasoning:**\\

[Your reasoning structure and step-by-step solution here]\\

**Final Answer:** [Correct Option Letter]

\end{minipage}}
\caption{Self-Discover Prompt Template for multiple-choice questions}
\label{fig:prompt_mcq_self_discover}
\end{figure}

\clearpage

\section{Essay Question Prompt Templates}
\label{sec:appendix_essay_prompts}

\subsection{Zero-Shot Essay Generation Prompt}
\label{subsec:essay_zero_shot}

\begin{figure}[h!]
\centering
\fbox{\begin{minipage}{0.95\textwidth}
\small
\small\ttfamily

**Topic:** \\
\{folder\}\\[0.5em]

**Vignette:**\\
\{vignette\}\\[0.5em]

**Question:**\\
\{question\}\\[0.5em]

**Instructions:**
Please provide a comprehensive, well-structured essay answer to the question above, based on the provided vignette and topic. Ensure your answer is clear, concise, and directly addresses all parts of the question. Show any calculations if the question requires them.\\

**Answer:**

\end{minipage}}
\caption{Zero-shot Prompt Template for essay generation.}
\label{fig:prompt_essay_default}
\end{figure}

\clearpage

\subsection{Chain-of-Thought Essay Generation Prompt}
\label{subsec:essay_cot}

\begin{figure}[h!]
\centering
\fbox{\begin{minipage}{0.95\textwidth}
\footnotesize
\small\ttfamily

**Topic:** \\
\{folder\}\\[0.5em]

**Vignette:**\\
\{vignette\}\\[0.5em]

**Question:**\\
\{question\}\\[0.5em]

**Instructions for Chain-of-Thought Essay Construction:**

You are a Chartered Financial Analyst (CFA) charterholder. Your task is to construct a comprehensive, well-structured essay answer to the question above. Follow these steps carefully:\\

1.  **Understand the Core Request:** Briefly rephrase the main objective of the question in your own words. What key information or analysis is being sought?\\

2.  **Identify Key Information \& Concepts:** Based on the vignette, topic, and question, list the critical pieces of information, CFA curriculum concepts, formulas, or analytical frameworks that will be relevant to constructing your answer.\\

3.  **Outline Your Essay Structure:** Before writing, create a brief bullet-point outline of how you will structure your essay. This should include the main sections or arguments you will present.\\

4.  **Step-by-Step Elaboration:** Following your outline, elaborate on each point.

    *   If calculations are needed, show your work clearly, explaining each step and the components of any formulas used.
    *   If discussing concepts, define them and explain their relevance to the question.
    *   Ensure your reasoning is logical and directly supported by the information in the vignette where applicable.\\

5.  **Synthesize and Conclude:** Briefly summarize your main points and provide a concluding statement that directly answers the question.\\

6.  **Review (Self-Correction):** Quickly review your drafted essay for clarity, accuracy, completeness, and conciseness. Ensure it directly addresses all parts of the question.
**Begin Essay Construction (following the steps above):**\\

[Your detailed, step-by-step constructed essay answer here]\\

\end{minipage}}
\caption{Chain-of-Thought Prompt Template for essay generation.}
\label{fig:prompt_essay_cot}
\end{figure}

\clearpage

\subsection{Self-Discover Essay Generation Prompt}
\label{subsec:essay_self_discover}

\begin{figure}[h!]
\centering
\fbox{\begin{minipage}{0.95\textwidth}
\footnotesize
\small\ttfamily

**Task:** Construct a comprehensive essay answer to the following question by first devising a reasoning structure using the Self-Discover method.\\

**Topic:** \\
\{folder\}\\[0.5em]

**Context/Vignette:**\\
\{vignette\}\\[0.5em]

**Question:**\\
\{question\}\\[0.5em]

**Self-Discover Essay Construction Process:**\\

**1. Select Reasoning Modules for Essay Construction:**
   Identify and list the core reasoning modules or types of thinking needed to construct a thorough essay for this specific question. Examples:
   
   -   **Problem Deconstruction:** Breaking down the question into smaller, manageable parts.
   -   **Information Extraction:** Identifying key facts, data, and constraints from the vignette.
   -   **Concept Application:** Determining relevant financial theories, models, or formulas (e.g., valuation methods, risk analysis, portfolio management principles).
   -   **Calculation \& Quantitative Analysis:** Performing necessary calculations and interpreting their results.
   -   **Qualitative Analysis \& Argumentation:** Developing logical arguments, discussing implications, and justifying conclusions.
   -   **Structuring \& Synthesis:** Organizing the analysis into a coherent essay structure.\\

**2. Adapt Modules to the Essay Task:**

   For each selected module, briefly explain how it specifically applies to constructing the essay for this question. Outline the key steps or considerations for each module in relation to the essay.
   For example:
   -   *Problem Deconstruction:* What are the sub-questions or core components the essay must address?
   -   *Information Extraction:* What specific data points from the vignette are crucial for each part of the essay?
   -   *Concept Application:* Which specific CFA curriculum concepts will form the backbone of the analysis? How will they be linked?\\

**3. Outline the Essay Structure (based on adapted modules):**

   Develop a high-level outline for your essay. This should detail the main sections (e.g., Introduction, Analysis of Factor X, Calculation of Y, Discussion of Implications, Conclusion) and the key points to be covered in each.\\

**4. Implement Reasoning and Write the Essay:**

   Execute the plan outlined above. Write the essay step-by-step, following your structured outline. Ensure you:
   -   Clearly explain your reasoning for each step or argument.
   -   Show all calculations if required, detailing the inputs and formulas.
   -   Integrate information from the vignette and the topic appropriately.
   -   Maintain a logical flow and clear, professional language suitable for a CFA context.\\

**Begin Essay Construction (following the Self-Discover process):**\\

[Model generates its reasoning structure (Modules, Adaptation, Outline) and then the full essay here]\\

**Full Essay Answer:**\\

[The complete essay answer should be placed here]\\

\end{minipage}}
\caption{Self-Discover Prompt Template for essay generation.}
\label{fig:prompt_essay_self_discover}
\end{figure}

\clearpage


\section{Grading Prompt Template}
\label{sec:appendix_grading_prompts}
\subsection{LLM-as-Judge Prompt}
\begin{figure}[h!]
\centering
\fbox{\begin{minipage}{0.95\textwidth}
\footnotesize
\small\ttfamily
You are a strict CFA Level III examination grader. You must follow the provided grading details EXACTLY - no partial credit beyond what is explicitly specified.\\

**CRITICAL GRADING REQUIREMENTS:**\\

- You MUST return ONLY a single integer between \{min\_score\} and \{max\_score\}\\[0.5em]

- Maximum possible score: \{max\_score\}\\[0.5em]

- Minimum possible score: \{min\_score\}\\[0.5em]

- NEVER exceed the maximum score under any circumstances
- Follow the grading details below EXACTLY - they specify the ONLY ways to earn points \\

**STRICT GRADING PROTOCOL:**

- Award points ONLY if the grading details criteria are met EXACTLY as specified

- Do NOT award partial credit unless explicitly mentioned in the grading details

- Do NOT give points for "close enough" answers - criteria must be met precisely

- Do NOT award points for effort, methodology, or partial understanding unless the grading details specify this

- If the grading details say "2 points for X", the student must demonstrate X completely to get those 2 points

- If multiple criteria exist (e.g., "2 points for A, 2 points for B"), each must be met independently

**QUESTION ASKED:**\\
\{question\}\\[0.5em]

**CONTEXT/VIGNETTE (if applicable):**\\
\{vignette\}\\[0.5em]

**GRADING DETAILS (FOLLOW EXACTLY):**\\
\{answer\_grading\_details\}\\[0.5em]

**CORRECT ANSWER/EXPLANATION (for reference only):**\\
\{correct\_answer\}\\[0.5em]

**STUDENT'S ANSWER:**\\
\{generated\_answer\}\\[0.5em]

**RESPONSE:** Return only the integer score based on strict adherence to the grading details above.
\end{minipage}}
\caption{GPT-4.1 Grading Prompt Template for essay grading}
\label{fig:prompt_grading_strict}
\end{figure}

\clearpage

\subsection{Self-Grade Prompt}
\label{subsec:grading_self}
\begin{figure}[h!]
\centering
\fbox{\begin{minipage}{0.95\textwidth}
\small\ttfamily

You are an expert CFA exam grader. Your task is to evaluate a generated answer to a CFA essay question based on the provided original question and a reference model answer.

You must provide a numerical score from 1 to 10, where 1 is very poor and 10 is excellent.
You must also provide a brief justification for your score, highlighting the strengths and weaknesses of the generated answer, especially in comparison to the reference model answer.

\{\\
\phantom{~~}"score": \textless integer\_score\_from\_1\_to\_10\textgreater,\\
\phantom{~~}"justification": "\textless your\_brief\_justification\_here\textgreater"\\
\}\\[0.5em]

**USER PROMPT:**\\
Original Question:\\
\{question\}\\[0.5em]

Reference Model Answer (for your guidance as a grader):\\
\{correct\_answer\}\\[0.5em]

Generated Answer to Evaluate:\\
\{generated\_answer\}\\[0.5em]

Please provide your evaluation in the specified JSON format.
\end{minipage}}
\caption{Self-Grade prompt template for holistic scoring.}
\label{fig:prompt_grading_self}
\end{figure}

\clearpage


\end{document}